\documentclass[conference]{IEEEtran}
\IEEEoverridecommandlockouts
\usepackage{cite}
\usepackage{amsmath,amssymb,amsfonts}
\usepackage{algorithmic}
\usepackage{graphicx}
\usepackage{textcomp}
\usepackage{xcolor}
\usepackage{multirow}
\usepackage{hyperref}
\def\BibTeX{{\rm B\kern-.05em{\sc i\kern-.025em b}\kern-.08em
    T\kern-.1667em\lower.7ex\hbox{E}\kern-.125emX}}
\begin{document}

\title{Fast-SEnSeI: Lightweight Sensor-Independent Cloud Masking for On-board Multispectral Sensors}

\author{
\IEEEauthorblockN{1\textsuperscript{st} Jan Kněžík}
\IEEEauthorblockA{\textit{Zaitra s.r.o.} \\
Brno, Czech Republic \\
jan.knezik@zaitra.io}
\and
\IEEEauthorblockN{2\textsuperscript{st} Jonáš Herec}
\IEEEauthorblockA{\textit{Zaitra s.r.o.} \\
Brno, Czech Republic \\
jonas.herec@zaitra.io}
\and
\IEEEauthorblockN{3\textsuperscript{st} Rado Pitoňák}
\IEEEauthorblockA{\textit{Zaitra s.r.o.} \\
Brno, Czech Republic \\
rado.pitonak@zaitra.io}
}

\maketitle

\begin{abstract}
Cloud segmentation is a critical preprocessing step for many Earth observation tasks, yet most models are tightly coupled to specific sensor configurations and rely on ground-based processing. In this work, we propose Fast-SEnSeI, a lightweight, sensor-independent encoder module that enables flexible, on-board cloud segmentation across multispectral sensors with varying band configurations. Building upon SEnSeI-v2, Fast-SEnSeI integrates an improved spectral descriptor, lightweight architecture, and robust padding-band handling. It accepts arbitrary combinations of spectral bands and their wavelengths, producing fixed-size feature maps that feed into a compact, quantized segmentation model based on a modified U-Net. The module runs efficiently on embedded CPUs using Apache TVM, while the segmentation model is deployed on FPGA, forming a CPU–FPGA hybrid pipeline suitable for space-qualified hardware. Evaluations on Sentinel-2 and Landsat 8 datasets demonstrate accurate segmentation across diverse input configurations.
\end{abstract}


\section{Introduction}
As the volume of satellite imagery captured in orbit continues to grow, the traditional paradigm of ground-based data processing is reaching its limits. Downlink bottlenecks, limited bandwidth, and the need for timely data products have driven the development of on-board artificial intelligence (AI) capabilities \cite{onboard1}, \cite{onboard2}, \cite{onboard3}, \cite{onboard4}. By moving parts of the processing pipeline directly onto the satellite, it becomes possible to filter, analyze, and prioritize data before transmission, enhancing mission efficiency and enabling new forms of real-time decision-making.

A particularly promising application of on-board AI is cloud segmentation. Optical Earth observation instruments frequently capture scenes that are partially or entirely obscured by clouds, rendering the data unsuitable for many downstream tasks. It is estimated that clouds cover around 70\% of the Earth’s surface at any time \cite{clouds}, meaning a large portion of optical imagery is either unusable or of limited value. The ability to detect clouds in orbit and assess acquisition quality in real-time allows satellites to optimize what is downlinked, either by discarding heavily clouded scenes or prioritizing those with clear-sky observations. This not only conserves bandwidth but also accelerates the delivery of high-quality data to end users.

However, on-board processing must operate within the strict limitations of embedded satellite hardware, where power availability is limited and computational resources are modest. To address this, a variety of edge platforms have been explored for deploying neural networks in orbit, including general-purpose CPUs, vision processing units (VPUs), GPUs, and field-programmable gate arrays (FPGAs). Each presents a different trade-off in terms of performance, flexibility, power efficiency, and space-readiness.

Beyond hardware efficiency, another critical consideration is sensor generalization. Most existing deep learning models in remote sensing are tightly coupled to the spectral characteristics of the sensor on which they were trained. Differences in the number, placement, and width of spectral bands between sensors pose a significant challenge to model reusability, cross-sensor deployment, and the ability to train on diverse, heterogeneous datasets. Sensor independence addresses this issue by enabling a model to ingest and process data from sensors with different spectral profiles. Rather than requiring perfect band alignment, a sensor-independent model can learn a shared representation space from arbitrary combinations of bands, provided these combinations fall within the domain seen during training. Approaches like SEnSeI-v1 \cite{senseiv1} and its successor SEnSeI-v2 \cite{senseiv2} have demonstrated how neural network architectures can abstract over sensor-specific characteristics and produce fixed-size feature maps compatible with downstream deep learning models.

In this work, we present Fast-SEnSeI, a lightweight, sensor-independent module designed specifically for on-board inference. Fast-SEnSeI takes as input a multispectral image along with its corresponding band wavelengths and produces a fixed-depth feature map suitable for downstream segmentation. The module runs entirely on a CPU using a Apache TVM compiled static graph. Its output is passed to a lightweight segmentation model deployed on an FPGA, which classifies each pixel into one of three categories: Clear, Thin Cloud, or Thick Cloud. This CPU-FPGA hybrid design is tailored for space-grade platforms such as the Zynq 7000 and Ultrascale+ families.

\section{Related Work}
Fast-SEnSeI builds upon the Spectral Encoder for Sensor Independence family, specifically the original SEnSeI-v1 and its successor SEnSeI-v2. These modules share the goal of enabling sensor-independent processing of multispectral satellite imagery by producing a fixed-format, spectrally informed representation for downstream deep learning models.

The core design of both SEnSeI-v1 and SEnSeI-v2 satisfies four key requirements for sensor independence:
\begin{itemize}
\item Flexible Input: They accept any number of spectral bands during inference.
\item Fixed Output: They produce a consistent number of channels as output, enabling compatibility with conventional CNN-based models.
\item Spectral Encoding: They encode the spectral characteristics of each band, allowing the subsequent model to reason about reflectance values in their appropriate spectral context.
\item Preserved Spatial Structure: All spatial information from the original input is retained, allowing pixel-wise tasks like segmentation to be performed without loss of resolution.
\end{itemize}

In SEnSeI-v1, the input consists of both the multispectral image and metadata describing each band’s spectral characteristics, specifically its minimum, central, and maximum wavelengths. The network processes each band’s descriptor vector independently through fully connected layers, and fuses cross-band information using a Permutation block, which constructs all pairwise combinations of features, processes them jointly, and pools the result to provide one refined feature vector per band. These spectral feature vectors are then spatially combined with the image by multiplying each feature vector with the corresponding input band, resulting in a set of spatial feature maps. These maps are subsequently averaged to produce a unified feature representation that preserves the original spatial dimensions while maintaining a fixed output depth, independent of the number of input bands.

SEnSeI-v2 builds on the original design by improving training efficiency, extending model capabilities, and increasing segmentation accuracy. Training is simplified by removing the need for pretraining the SEnSeI module in isolation (i.e., before attaching a downstream task-specific model), making the architecture easier to integrate into conventional pipelines. The architecture also introduces support for multimodal inputs, such as SAR and DEM. Accuracy gains are achieved through several architectural modifications. First, the minimum and maximum wavelength of each band are transformed using sinusoidal spectral encodings, inspired by transformer positional encodings, which capture spectral context. These transformed descriptors are then extended with band-level reflectance statistics, enabling the model to better capture scene-level variations. In place of the Permutation block, SEnSeI-v2 employs a transformer-based attention module for cross-band feature fusion. Finally, rather than simply multiplying reflectance values by spectral features, SEnSeI-v2 applies a learnable sinusoidal embedding mechanism, which allows the model to control how reflectance values are mapped into the final output tensor.

Experimental results from the original work show that SEnSeI-v2 achieves state-of-the-art performance on cloud segmentation tasks. It outperforms SEnSeI-v1 while remaining competitive with non-sensor-independent models. When integrated with architectures like DeepLabv3+ and SegFormer and trained on the Sentinel-2 CloudSEN12 dataset, SEnSeI-v2 consistently maintains high accuracy. Moreover, SegFormer models using SEnSeI-v2 and trained on multi-sensor data (Sentinel-2, Landsat 7, Landsat 8, PeruSat-1) outperform models trained solely on Landsat 8, even when evaluated on a distinct Landsat 8/9 test set, highlighting the value of sensor independence in improving generalization across platforms.

In this work, we adapt components from SEnSeI-v1 and SEnSeI-v2 and introduce additional changes to enable efficient CPU deployment with Apache TVM. For segmentation, we use a simpler network architecture than in the SEnSeI papers, which enables FPGA deployment and real-time inference on-board satellites.

\section{Methods}
\subsection{Fast-SEnSeI}
Fast-SEnSeI introduces four targeted changes: (a) improved descriptor design, (b) SEnSeI-v2 body with a SEnSeI-v1 output block, (c) reduced descriptor and layer widths, and (d) improved padding-band handling.\newline

\paragraph{Descriptor design}
SEnSeI-v2 uses a sinusoidal ``spectral encoding'' together with a logarithmic wavelength normalization, which together skew the set of sinusoid frequencies and collapse variation in many encoding dimensions. More concretely, for wavelength \(\lambda\) in nanometers it normalizes it as
\[
\lambda_{\mathrm{norm}} = \log_{10}(\lambda-300)-2,
\]
then computes
\[
x_i = \frac{\lambda_{\mathrm{norm}}}{10000^{\,2i/D}}, \qquad i = 0,\ldots,D-1,
\]
and constructs the encoding as
\[
\big[\sin x_0,\; \cos x_1,\; \sin x_2,\; \cos x_3,\; \ldots,\; \sin x_{D-2},\; \cos x_{D-1}\big].
\]
Thus, within each \(\sin\)–\(\cos\) pair, \(\lambda_{\mathrm{norm}}\) is divided by different denominators, so the pair does not share a common frequency. This approach also pushes the largest denominator toward \(10000^{2}\) rather than remaining below \(10000^{1}\). Together with the compression from the logarithmic normalization into a narrow numeric interval, this makes several encoding dimensions vary only negligibly over the wavelength range and therefore limits the expressiveness of the encoding.

We fix these issues by adopting the standard Transformer formulation \cite{transformer} with \(D/2\) shared frequencies and a simple normalization that avoids compression. We use a shift-only normalization without scaling, since we restrict our study to VNIR bands. For wavelengths \(\lambda\) in nanometers we set
\[
\lambda_{\mathrm{norm}} = \lambda - 400,
\]
then define
\[
y_j = \frac{\lambda_{\mathrm{norm}}}{10000^{\,2j/D}}, \qquad j = 0,\ldots,\tfrac{D}{2}-1,
\]
and constructs the encoding as
\[
\big[\sin y_0,\; \cos y_0,\; \sin y_1,\; \cos y_1,\; \ldots,\; \sin y_{D/2-1},\; \cos y_{D/2-1}\big].
\]
With this definition, each encoding dimension spans many more sinusoidal cycles across the wavelength range of the bands, which enhances the distinctiveness and sensitivity of the encoding.

In SEnSeI-v2, the entire descriptor consists of the spectral encoding for minimum and maximum wavelength, binary flags for non-optical modalities (e.g., SAR, DEM, thermal), and band-level reflectance statistics. We remove the binary flags from the descriptor, since we focus on optical VNIR bands. The band-level reflectance statistics were originally five percentiles (1\%, 10\%, 50\%, 90\%, 99\%), which were not compatible with TVM-based model compilation. We therefore replace them with four summary statistics (minimum, maximum, mean, standard deviation), which are supported.\newline

\paragraph{Architecture}
The architecture combines the SEnSeI-v2 body with the SEnSeI-v1 output block. The body operates on low-dimensional wavelength descriptors, while the output block operates on image-shaped feature maps and therefore dominates computation. To keep this stage efficient, we adopt the SEnSeI-v1 Band Multiplication block instead of the Band Embedding block with learnable sinusoidal encoding used in SEnSeI-v2, as the latter reduces throughput by about 3.5×. We also reduce the number of output feature maps from 32 to 4, since higher dimensionality decreases throughput by more than 4×. The experiments supporting these measurements are presented in Subsection~\ref{section:modifications_results}.\newline

\paragraph{Model size reduction}
To reduce model size, we shrink the spectral encoding from 64 to 32 dimensions. We encode each band by concatenating the encodings of its minimum and maximum wavelengths (16 + 16 = 32 dims), and append four band-level reflectance statistics (minimum, maximum, mean, standard deviation), providing a 36-dimensional per-band descriptor. A per-band multilayer perceptron (MLP) expands these descriptors 36\textrightarrow 48\textrightarrow 64, producing a 64-dimensional token per band; the sequence length equals the number of valid bands. The Transformer encoder then processes these tokens with multi-head self-attention and feed-forward network that expands the representation from 64 to 256 dimensions before contracting back to 64, maintaining the standard 4× expansion ratio commonly used in Transformer models \cite{transformer}. Finally, a per-band MLP contracts the representation 64\textrightarrow 32\textrightarrow 16\textrightarrow \(C\), where \(C\) is the number of output feature maps. This expand–then–contract pattern was not present in SEnSeI-v2 and provides a more compact network while maintaining accuracy in our experiments. Overall, the number of trainable parameters decreased from 355k to 117k.\newline

\paragraph{Padding-band handling}
During training, the number and combination of input bands per sample is deliberately randomized to make the model robust to arbitrary band subsets at inference. Since batches require uniform shapes, missing bands are represented by padding-bands and their descriptors by padding-descriptors. In SEnSeI-v1, padding-bands are set to -0.5, which, when added with +0.5 in the Band Multiplication block, provides zeros; thus padding-bands do not contribute when feature maps are summed, regardless of the values in processed padding-descriptors. However, because we compute the mean of band feature maps to avoid scale growth with band count, zero-valued maps would still bias the average if they were included in the divisor. We therefore divide only by the number of real bands. 

Moreover, because Fast-SEnSeI uses self-attention layers, as in SEnSeI-v2, the padding-descriptors could lead to unintended cross-band interference. To prevent this, we add an attention padding mask so queries, keys, and values corresponding to padding-descriptors are ignored during self-attention.

To quantify the effect of padding, we compare three variants:
\begin{itemize}
    \item Level 1: Averages over all bands (including padding).
    \item Level 2: Averages only over real bands.
    \item Level 3: Equals Level 2 plus an attention padding mask.
\end{itemize}

\subsection{Lightweight Segmentation Network}
The segmentation model is designed for deployment on a resource-constrained FPGA. We adopt a modified U-Net–like architecture \cite{unet}, with attention to balancing the number of operations across layers to optimize inference speed and hardware utilization on the target device. The architecture consists of four downsampling stages, followed by a bottleneck, and four upsampling stages, with each stage containing two convolutional layers with 3×3 kernels. Channel widths follow a lightweight symmetric schedule: we start with 8 channels and double at each encoder stage and in the bottleneck, then halve at each decoder stage. We introduce a few small deviations from this rule to balance FLOPs.

Several architectural variations were evaluated during development, including the use of transposed convolutions versus nearest-neighbor interpolation for upsampling, as well as different skip connection strategies. While individual changes resulted in only marginal variations in segmentation performance (±0.5 mIoU), the best configuration used nearest-neighbor interpolation without skip connections. Notably, the removal of skip connections did not result in a performance drop, which is consistent with previous findings \cite{skip} suggesting that while skip connections are useful for preserving high-frequency details in the decoder, they are less effective for cloud segmentation since clouds typically lack such high frequency information.

\subsection{Heterogeneous Deployment}
Fast-SEnSeI's architectural complexity makes it unsuitable for FPGA deployment. However, due to its efficient execution characteristics, it can run effectively on a CPU. In the proposed system, we leverage this by executing the sensor-independent module on a CPU and feed the resulting image-shaped feature maps to the FPGA-accelerated segmentation network. 

For compilation and deployment of Fast-SEnSeI, we use Apache TVM, an open-source machine learning compiler that produces highly optimized static graphs. Only the compiled module and a lightweight runtime ($\approx 2$~MB) need to be stored or uplinked to the satellite, assuming a C++ runtime environment is available.

To meet FPGA resource budgets, the Lightweight Segmentation Network is quantized with Quantization-Aware Training (QAT) using Brevitas \cite{brevitas}. The first and last convolutional layers use 8-bit weights, all intermediate layers use 4-bit weights, and activations are quantized to 4 bits. This combination of weight and activation quantization results in a compact, hardware-efficient model with only minimal degradation in segmentation accuracy. The quantized Brevitas model is exported to QONNX and lowered with FINN \cite{finn} into a streaming dataflow accelerator using HLS. The high-level design (C++) is synthesized to RTL with Vitis HLS, packaged as a Vivado IP core, integrated into the top-level design, and implemented to produce the final FPGA bitstream. During deployment, we tune per-layer folding and parallelism, specifically the number of processing elements (PEs; output-channel parallelism) and SIMD lanes per PE (input-channel parallelism), to meet the cycle budget and device resource constraints.

\section{Datasets}
We used three publicly available cloud masking datasets: L8-Biome (Landsat 8), CloudSEN12+, and KappaSet (Sentinel-2). Each dataset includes pixel-level annotations for different cloud types and was preprocessed to produce standardized 512×512 tiles suitable for training and evaluation of the Fast-SEnSeI module and the FPGA-based segmentation model.

\subsection{L8-Biome}
The L8-Biome dataset \cite{l8biome} contains full Level-1T Landsat 8 scenes of approximately 8000×8000 pixels. For each scene, cloud masks were produced through manual annotation at 30~m resolution, distinguishing four classes: Clear, Cloud shadow, Thin cloud, and Thick cloud.

The original scenes are georeferenced with geodetic north aligned at the top, which introduces no-data regions due to satellite tilt. To maximize usable image area, each scene was rotated by the inverse of the tilt angle and subsequently cropped. The same transformation was applied to the corresponding cloud masks. The resulting scenes were then split into non-overlapping 512×512 pixel tiles, providing a total of 13,686 tiles.

Radiometric calibration was performed by converting digital numbers (DN) to Top-of-Atmosphere (TOA) reflectance using the published scaling factors:
\[
\text{reflectance} = \text{DN} \times 2 \times 10^{-5} - 0.1
\]
The resulting TOA reflectance values typically fall within the range 0–1, with occasional values above 1 due to effects such as specular reflection or off-nadir surface geometry.

\subsection{CloudSEN12+}
The CloudSEN12+ dataset \cite{cloudsen} consists of 10,000 Sentinel-2 image patches, each measuring 509×509 pixels. Cloud masks were manually created at 10 m resolution and define four categories: Clear, Cloud shadow, Thin cloud, and Thick cloud. To ensure consistency across inputs, the original 20~m and 60~m Sentinel-2 bands were resampled to 10~m. The patches are drawn from multiple locations across all continents except Antarctica. No additional tiling was required, as the patches are already provided with padding to 512×512 pixels in the original dataset.

All pixel values were converted to Top-of-Atmosphere (TOA) reflectance by dividing the Level-1C digital numbers (DN) by 10,000. The resulting values typically lie within the range 0–1, with occasional oversaturated outliers exceeding this range.

\subsection{KappaSet}
The KappaSet \cite{kappaset} dataset includes 9,251 Sentinel-2 image tiles, each measuring 512×512 pixels, extracted from 1,038 Level-1C scenes. Manual cloud annotations are provided at 10~m resolution, covering five categories: Undefined (unsure), Clear, Cloud shadow, Thin cloud, and Thick cloud. Bands originally captured at 20~m or 60~m resolution were resampled to match the 10~m scale. The selection ensures broad geographical and seasonal diversity.

The KappaSet dataset was normalized in the same way as CloudSEN12+, by dividing DN values by 10,000 to obtain TOA reflectance in the 0–1 range, with rare oversaturated values exceeding this interval.

\section{Model Training}
All models were trained using only bands within the visible and near-infrared (VNIR) spectral range. The specific Sentinel-2 and Landsat 8 band sets are listed in Table \ref{tab:bands}. We use the shorthand \emph{All bands} for the complete Sentinel-2 set and \emph{Common bands} for the intersection of Sentinel-2 and Landsat 8, which equals the full Landsat 8 set. Although the two sensors only share five common bands, the sensor-independent nature of the model allows it to learn from all available bands across both datasets.

\begin{table}[htbp]
\caption{VNIR band sets used for training and evaluation.}
\label{tab:bands}
\centering
\begin{tabular}{|l|p{0.65\linewidth}|}
\hline
\textbf{Set} & \textbf{Bands} \\
\hline
All bands \rule{0pt}{5ex} & \multirow{2}{*}{\parbox{0.95\linewidth}{Coastal aerosol, Blue, Green, Red, Red Edge 1, Red Edge 2, Red Edge 3, NIR, Narrow NIR, Water vapor}} \\
\cline{1-1}
Sentinel-2 \rule{0pt}{5ex} & \\
\hline
Common bands \rule{0pt}{3.0ex} & \multirow{2}{*}{Coastal aerosol, Blue, Green, Red, NIR} \\
\cline{1-1}
Landsat 8 \rule{0pt}{3.0ex} & \\
\hline
\end{tabular}
\end{table}

A random subset of bands was selected for each sample during training, between from 1 and the number available for that sensor (up to 10 for Sentinel-2 and up to 5 for Landsat 8). This encourages the model to handle arbitrary band combinations at inference time, simulating the conditions required for true sensor independence.

The dataset was split into 60\% training, 20\% validation, and 20\% test subsets. To enhance generalization, basic data augmentation techniques were applied, including random 90° rotations and horizontal/vertical flips.

Training was conducted using three segmentation classes: Clear, Thin cloud, and Thick cloud. The Cloud shadow class from the original annotations was merged into Clear to simplify the task and reduce class imbalance.

Both the segmentation model and Fast-SEnSeI were trained on a single NVIDIA GeForce RTX 3090 for 50 epochs using the AdamW optimizer with a batch size of 64. The initial learning rate was set to 0.0005, with a weight decay of 0.005. The learning rate was linearly increased from 10\% of its base value to the full value during the first three epochs, and then gradually decreased to 20\% of the base value by the final epoch using cosine annealing. Training used pixel-wise cross-entropy loss, and the final model was selected as the checkpoint from the epoch with the highest validation mIoU.

To reduce training time, the segmentation models were trained in their full-precision (non-quantized) form. Quantization using QAT was applied only to the few final models to measure inference speed on hardware and to compare quantized versus non-quantized evaluation performance. During quantization, reflectance values normalized to the 0–1 range were rescaled to 0–255, clipped, rounded, and finally represented as 8-bit unsigned integers.

\section{Results}
\begin{figure}[htbp]
  \centering
  \includegraphics[width=0.48\textwidth]{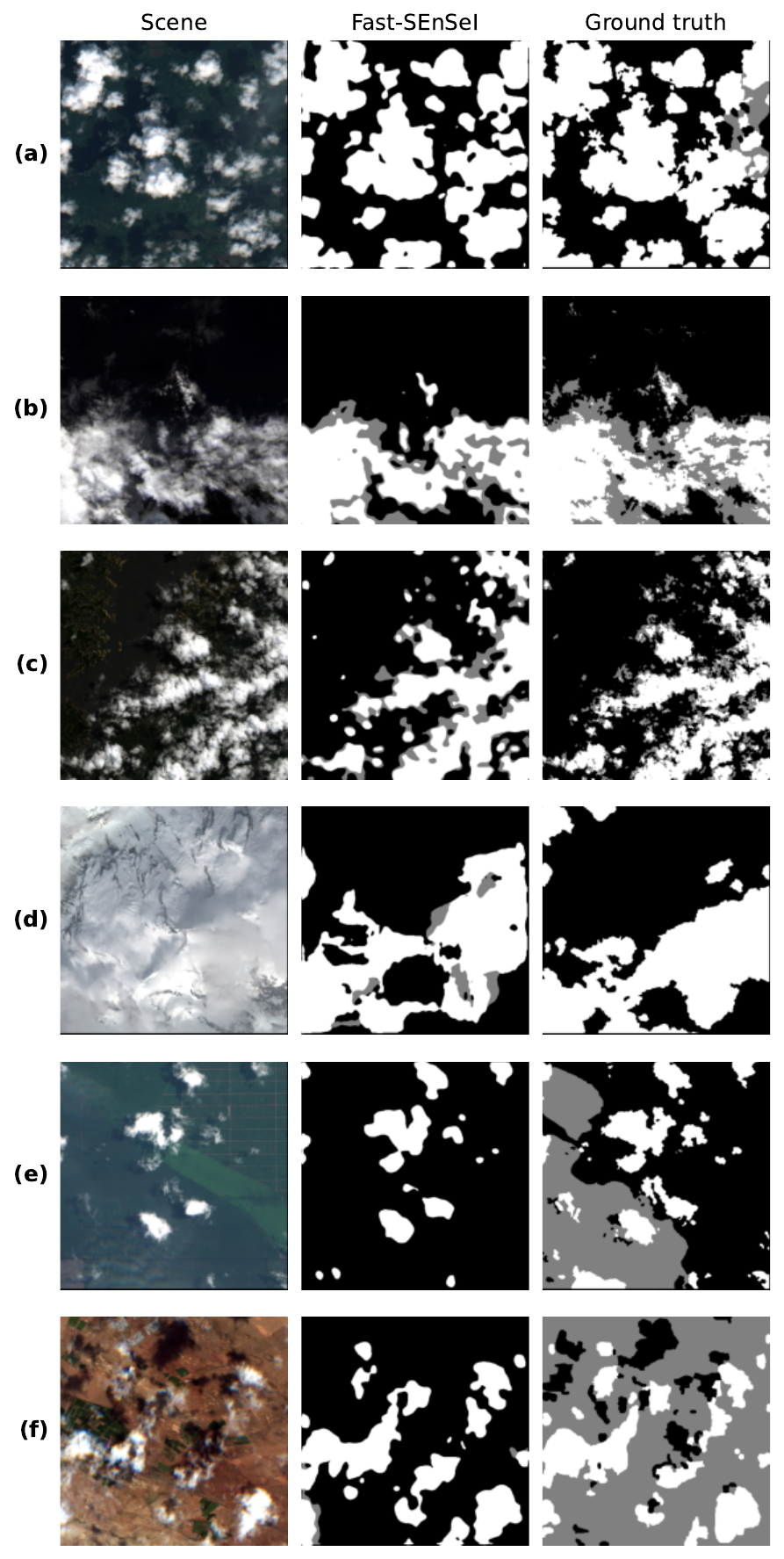}
  \caption{Fast-SEnSeI predictions on the test sets of all datasets, obtained using Common bands.
   The shown scenes were hand-picked to highlight specific properties of the model and data. 
   Examples (a)-(c) illustrate a limitation of the model: in regions with dense and fine-grained cloud structures, the predictions tend to appear more smoothed compared to the ground truth masks.
   Example (d) demonstrates that the model transfers well to snow-covered regions.
   Examples (e) and (f) show cases where thin clouds are missed. However, these thin clouds are not (or only faintly) visible in the RGB imagery, suggesting that either larger spatial context beyond the $512 \times 512$ patch or additional spectral information may be necessary to identify them.
   }
  \label{fig:fast_sensei_examples}
\end{figure}

We evaluated the proposed sensor-independent segmentation model using a series of controlled experiments to assess accuracy and hardware performance. Unless stated otherwise, all results are reported using Fast-SEnSeI, which incorporates an improved descriptor design (modified normalization and spectral encoding), TVM-compatible band-level reflectance statistics (minimum, maximum, mean, standard deviation), a Band Multiplication output block, reduced model size with 4 output feature maps, and padding-band handling level 3. Selected Fast-SEnSeI predictions illustrating special cases are shown in Fig.~\ref{fig:fast_sensei_examples}.

\subsection{Band Combinations}
To assess the ability of Fast-SEnSeI to handle different spectral input configurations, we trained it on all datasets with all available bands and evaluated it in two settings: all datasets (both Sentinel-2 and Landsat 8), and only Sentinel-2 datasets. As shown in Table \ref{tab:band_combinations}, the model performs reliably even with a single input band when trained on all datasets, and performance improves steadily as more bands are added.

For the Sentinel-2-only evaluations, we compared the commonly used RGB+NIR bands against the remaining six spectral bands. The RGB+NIR configuration performed slightly better, but the difference was small, showing that the other bands also provide valuable information. This is confirmed by the fact that the best results were achieved when all ten Sentinel-2 bands (RGB+NIR plus the remaining six) were used together.

\begin{table*}[htbp]
\caption{Fast-SEnSeI segmentation performance with different input band combinations. Results are grouped by dataset configuration. Metrics are reported separately for Clear, Thin cloud, and Thick cloud classes (Prec = Precision, Rec = Recall, IoU = Intersection over Union).}
\begin{center}
\begin{tabular}{|c|c|ccc|ccc|ccc|c|}
\hline
\multirow{2}{*}{\textbf{Datasets}}
 & \multirow{2}{*}{\textbf{Bands}}
 & \multicolumn{3}{c|}{\textbf{Clear}}
 & \multicolumn{3}{c|}{\textbf{Thick cloud}}
 & \multicolumn{3}{c|}{\textbf{Thin cloud}} 
 & \textbf{All classes} \\
\cline{3-12}
 & 
 & \textbf{Prec}
 & \textbf{Rec}
 & \textbf{IoU}
 & \textbf{Prec}
 & \textbf{Rec}
 & \textbf{IoU}
 & \textbf{Prec}
 & \textbf{Rec}
 & \textbf{IoU} 
 & \textbf{mIoU} \\
\hline
\multirow{3}{*}{\textbf{All}}
 & Red
 & 87.66 & 93.38 & 82.53
 & 85.25 & 87.51 & 76
 & 62.04 & 44.11 & 34.74 
 & 64.42 \\
 & RGB
 & 89.34 & 94.38 & 84.82
 & 85.66 & 90.45 & 78.56
 & 68.16 & 46.72 & 38.35 
 & 67.24 \\
 & RGB+NIR
 & 89.49 & 94.29 & 84.89
 & 87.54 & 90.98 & 80.54
 & 68.35 & 49.75 & 40.44 
 & 68.62 \\
 & Common bands
 & 90.05 & 94.17 & \textbf{85.29}
 & 88.13 & 90.8 & \textbf{80.9}
 & 67.84 & 52.54 & \textbf{42.06} 
 & \textbf{69.42} \\
\hline
\multirow{3}{*}{\textbf{Sentinel-2}}
 & RGB+NIR
 & 86.09 & 93.72 & 81.39
 & 86.04 & 89.14 & \textbf{77.88}
 & 68.42 & 42.7 & 35.67 
 & 64.98 \\
 & All w/o RGB+NIR
 & 86.37 & 91.25 & 79.76
 & 86.25 & 85.17 & 74.99
 & 61.04 & 50.24 & 38.04 
 & 64.26 \\
 & All bands 
 & 87.55 & 92.61 & \textbf{81.83}
 & 86.14 & 88.37 & 77.38
 & 66.02 & 49.57 & \textbf{39.5} 
 & \textbf{66.24} \\
\hline
\end{tabular}
\label{tab:band_combinations}
\end{center}
\end{table*}

\subsection{Fast-SEnSeI modifications}
\label{section:modifications_results}
\begin{figure}[htbp]
  \centering
  \includegraphics[width=0.48\textwidth]{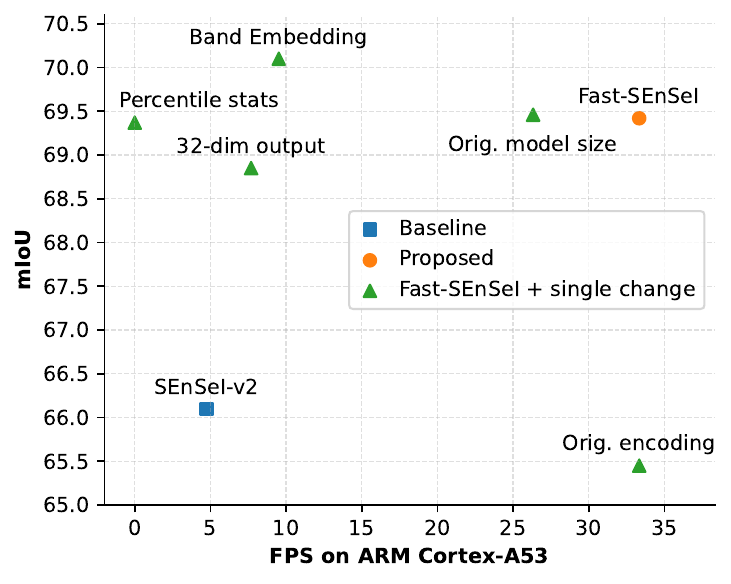}
  \caption{Comparison of inference speed (FPS) and segmentation accuracy (mIoU) between the baseline SEnSeI-v2, the proposed Fast-SEnSeI, and variants where a single component of Fast-SEnSeI is modified.}
  \label{fig:fps-miou-graph}
\end{figure}

To assess the efficiency of the proposed approach, we evaluated inference speed and segmentation accuracy for SEnSeI-v2, Fast-SEnSeI, and its single-change variants on the ARM Cortex-A53. The reported frames per second (FPS) correspond exclusively to the SEnSeI module, while the mean Intersection-over-Union (mIoU) is measured for the complete pipeline that combines SEnSeI with the segmentation model. All models were trained on the full set of datasets with all available bands. Evaluation was conducted on the same datasets, but restricted to Common bands, with mIoU reported across three classes: Clear, Thin cloud, and Thick cloud. The results are summarized in Fig.~\ref{fig:fps-miou-graph}.

Among the evaluated design choices, the \textbf{Percentile statistics} variant achieves nearly the same mIoU as the TVM-compatible band-level reflectance statistics (69.37 vs. 69.42), but it cannot be deployed with TVM and therefore appears at 0 FPS in the figure. Switching to the TVM-compatible statistics thus enables hardware deployment without loss in performance. The largest performance penalty in terms of speed arises from the \textbf{32-dim output} variant, where the module produces 32 output feature maps. This not only reduces FPS (33.33$\rightarrow$7.69) but would also increase the latency of the downstream segmentation model, as it must process a higher-dimensional input. The \textbf{Band Embedding} variant, which incorporates a learnable sinusoidal embedding block, also reduces FPS substantially (33.33$\rightarrow$9.52) but achieves the highest mIoU (70.10). However, the marginal mIoU gain over Fast-SEnSeI does not compensate for the significant decrease in throughput. In contrast, the \textbf{Orig. spectral encoding} leads to the largest drop in mIoU (65.45) despite running at the same speed as Fast-SEnSeI, confirming the effectiveness of the improved encoding design. Finally, the \textbf{Orig. model size} variant achieves nearly identical mIoU (69.46 vs. 69.42) but is slower (26.32 vs. 33.33 FPS) than Fast-SEnSeI. This means the reduction in model size comes at no mIoU cost and provides a gain in FPS, while also reducing the deployment footprint: the TVM-compiled binary on ARM Cortex-A53 is 1.3 MB for the reduced-size model compared to 2.0 MB for the original, with similar ratios observed on ARM Cortex-A9 (1.1 MB vs. 1.8 MB) and in hardware-agnostic ONNX format (0.5 MB vs. 1.3 MB).

The baseline \textbf{SEnSeI-v2} can be viewed as the combination of all single modifications: 32-dim output, Band Embedding, original model size, and original spectral encoding, together with Percentile statistics. To enable a fair runtime comparison, its FPS is reported with TVM-compatible statistics, while accuracy is evaluated with Percentile statistics. This compound design inherits the slowdown of the variants and suffers the mIoU degradation of the original spectral encoding, with only a modest mIoU improvement from Band Embedding. As a result, \textbf{Fast-SEnSeI} achieves both higher mIoU (69.42 vs. 66.1) and significantly higher throughput (33.33 vs. 4.76 FPS) than SEnSeI-v2, highlighting the effectiveness of the proposed modifications.

These results should also be interpreted with respect to class imbalance. The Thin cloud class is underrepresented in the datasets, and the model performs poorly on it, which lowers the overall mIoU. When the task is simplified to binary Cloud/Clear segmentation, Fast-SEnSeI reaches 83.4 mIoU, compared to 69.42 in the three-class setting.

\subsection{Padding-Band Handling Strategies}
\begin{figure}[htbp]
  \centering
  \includegraphics[width=0.48\textwidth]{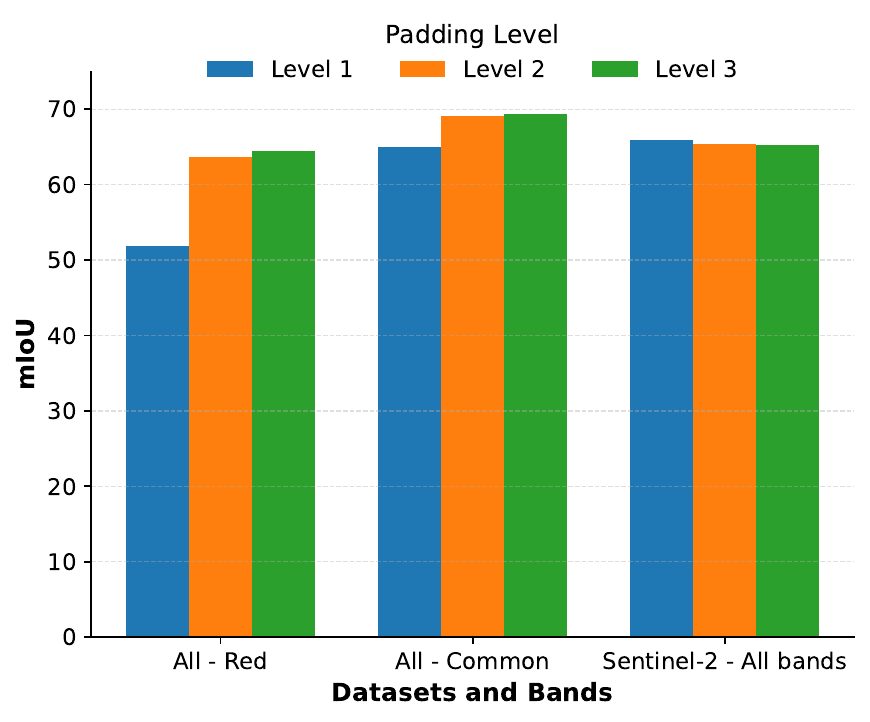}
  \caption{Comparison of mIoU for three padding-band handling strategies across different number of bands. Level 1 averages over all bands including padding, Level 2 averages only real bands, and Level 3 applies Level 2 with an attention padding mask.}
  \label{fig:padding-band-miou}
\end{figure}

To better understand the impact of padding-band handling, we trained three Fast-SEnSeI models on all datasets with all bands, each using a different padding-band handling level, and evaluated them across various dataset configurations, with the results summarized in Fig~\ref{fig:padding-band-miou}.

The largest effect is observed in the \textbf{All – Red} configuration, where all datasets are restricted to a single band. This is because whenever the model encountered only one real band during training, the remaining inputs were padding-bands. Their treatment therefore becomes particularly important to prevent the model from becoming dependent on them, since they are not used during inference. Specifically, Level 1, which averages across all bands including padded ones, performs poorly and achieves an mIoU of only 51.86. Moving to Level 2, which excludes padding from the averaging, leads to a dramatic improvement of almost twelve points, reaching 63.59. The addition of the attention padding mask in Level 3 provides a smaller further increase to 64.42.

The \textbf{All – Common} configuration, which uses the five spectral bands shared across all datasets, shows more moderate improvements. Here, Level 1 again lags behind with an mIoU of 64.99, while Level 2 recovers accuracy to 69.08. The effect of Level 3 is marginal, raising the score only slightly further to 69.42.

Finally, the \textbf{Sentinel-2 – All bands} configuration, which leverages all ten bands of Sentinel-2, shows almost no difference between the three strategies. With no padding involved, the results remain stable across levels, with mIoUs of 65.89, 65.33, and 65.24, respectively.

Overall, these findings demonstrate that padding-band handling is most critical when only a few spectral bands from the full set used during training are available at inference. When more bands are available, the role of padding diminishes, and all strategies converge to nearly identical results. Importantly, Level 3 doesn't reduce performance and is especially valuable in low-band settings, making it the safest and most robust option across different scenarios.

\subsection{Impact of Fast-SEnSeI and Quantization on Accuracy}
To evaluate the impact of Fast-SEnSeI and quantization of segmentation model on accuracy, we compared three variants on the Common bands of all datasets: Fast-SEnSeI with a non-quantized segmentation model (the default configuration used in previous experiments), Fast-SEnSeI with a quantized segmentation model, and a quantized segmentation model without Fast-SEnSeI. In both Fast-SEnSeI variants, the models were trained on all datasets with all available bands, whereas the model without Fast-SEnSeI was trained directly on the Common bands. As shown in Table \ref{tab:without_fast_sensei}, Fast-SEnSeI with quantization shows a small decrease in accuracy compared to its non-quantized counterpart. This drop is minor and represents a necessary compromise to enable deployment on FPGA hardware. The segmentation model without Fast-SEnSeI achieves higher metrics, as expected, but this comes at the cost of being restricted to the fixed common band set. The slight performance reduction observed with Fast-SEnSeI may also be due to the need for the compact segmentation model to generalize across varying input distributions. Although SEnSeI maps different band combinations into a common feature space, the resulting representations can still vary depending on the specific combination of bands, making the segmentation task more challenging than when training on a fixed band set.

\begin{table}[htbp]
\caption{IoU score evaluation of common bands on all datasets: Fast-SEnSeI with non-quantized and quantized segmentation models, and quantized segmentation model trained directly on the common bands.}
\label{tab:without_fast_sensei}
\centering
\begin{tabular}{|c|c|ccc|c|}
\hline
\textbf{Fast-SEnSeI} & \textbf{QAT} & \textbf{Clear} & \textbf{Thick cl.} & \textbf{Thin cl.} & \textbf{All (mIoU)} \\
\hline
Yes & No & 85.29 & 80.9 & 42.06 & 69.42 \\
Yes & Yes & 84.42 & 80.16 & 40.85 & 68.48 \\
No  & Yes & 86.31 & 81.21 & 48.02 & 71.85 \\
\hline
\end{tabular}
\end{table}

To showcase the restriction to the fixed band set of the segmentation model without Fast-SEnSeI, we performed an evaluation with band replacement, as shown in Table \ref{tab:without_fast_sensei_2}. It shows that when a single band was substituted with a spectrally different one not seen during training, performance degraded significantly. In the first example (Vapor+RGB+NIR), the Coastal aerosol band was replaced with Water vapor. This band has a similar 60 m GSD and 20 nm bandwidth, but it is located on the opposite side of the VNIR spectrum, with a central wavelength shift of 501 nm. This caused a major drop in all evaluation metrics. In the second example (Aero+RE3+GB+NIR), the Red band was replaced with Red Edge 3. This band has a smaller spectral shift of 115 nm, and performance decreased less, but still noticeably.

These results confirm that traditional models are highly sensitive to band configuration, unlike the model with Fast-SEnSeI shown in Table \ref{tab:band_combinations}.

\begin{table}[htbp]
\caption{Generalization test with IoU scores for a model without Fast-SEnSeI trained on the common bands, where individual bands are replaced with unseen alternatives.}
\label{tab:without_fast_sensei_2}
\centering
\footnotesize
\begin{tabular}{|c|c|ccc|}
\hline
\textbf{Datasets} & \textbf{Bands} & \textbf{Clear} & \textbf{Thick cl.} & \textbf{Thin cl.} \\
\hline
\multirow{3}{*}{Sentinel-2}
 & Common bands    & 84.65 & 78.88 & 49.66 \\
 & Vapor+RGB+NIR   & 42.43 & 16.19 & 16.71 \\
 & Aero+RE3+GB+NIR & 77.15 & 68.24 & 25.93 \\
\hline
\end{tabular}
\end{table}

\subsection{Inference Performance and Resource Utilization}
To evaluate real-time deployability, we measured Fast-SEnSeI’s inference latency and memory usage on ARM Cortex-A9 and Cortex-A53 processors using input configurations with 1, 5, and 10 spectral bands. The results, summarized in Table \ref{tab:cpu_measurements}, show that runtime increases sharply as the number of input channels grows, whereas RAM usage rises only moderately. This indicates that runtime is the primary limiting factor when scaling to higher input dimensionalities. Because the segmentation model always processes the fixed output size of Fast-SEnSeI, its runtime remains independent of the number of input bands. The choice of input channel count therefore represents a trade-off between runtime, as reported here, and accuracy, as reported in Table \ref{tab:band_combinations}.

\begin{table}[htbp]
\caption{Inference latency and resource usage of Fast-SEnSeI on ARM Cortex-A9 and Cortex-A53 CPUs, measured for varying numbers of input bands.}
\label{tab:cpu_measurements}
\centering
\begin{tabular}{|c|cc|cc|}
\hline
\multirow{2}{*}{\textbf{In ch.}} 
 & \multicolumn{2}{c|}{\textbf{ARM Cortex-A9}} 
 & \multicolumn{2}{c|}{\textbf{ARM Cortex-A53}} \\
\cline{2-5}
 & \textbf{Lat. (ms)} 
 & \textbf{RAM (MB)} 
 & \textbf{Lat.} 
 & \textbf{RAM} \\
\hline
 1 & 66 & 58 & 9.5 & 57 \\
 5 & 186 & 70 & 30 & 69 \\
 10 & 343 & 80 & 91 & 79 \\
\hline
\end{tabular}
\end{table}

In addition to CPU performance, we evaluated the inference time of the segmentation model on FPGA. The results are shown in Table \ref{tab:fpga_measurements}, which summarizes frame processing time and hardware resource utilization on two FPGA platforms. The folding configuration was chosen such that the design fits on the resource-constrained Zynq 7000 device. Although the Ultrascale+ provides a significantly larger resource budget, we employed the same folding configuration, resulting in identical frame processing time. While these additional resources could be exploited to reduce runtime through more aggressive folding, such optimization is not explored in this work.
Overall, these results show that the full Fast-SEnSeI pipeline, with 5-band input at 512×512 resolution, can be executed in approximately 30 ms (Fast-SEnSeI on ARM Cortex-A53) plus 324 ms (segmentation on FPGA).

\begin{table}[htbp]
\caption{Segmentation model performance and resource utilization on FPGA.}
\label{tab:fpga_measurements}
\centering
\begin{tabular}{|c|c|c|c|c|c|}
\hline
\textbf{FPGA} & \textbf{Latency (ms)} & \textbf{LUT} & \textbf{FF} & \textbf{BRAM} & \textbf{DSP} \\
\hline
Zynq 7000 & 324 & 37090 & 51087 & 93.5 & 211 \\
Ultrascale+ & 324 & 36828 & 50834 & 75 & 233 \\
\hline
\end{tabular}
\end{table}

\section{Conclusion}
This work presents Fast-SEnSeI, a lightweight, sensor-independent encoder designed for on-board cloud segmentation. The module converts arbitrary combinations of multispectral bands and their wavelengths into a fixed number of output channels, enabling compatibility with downstream models regardless of sensor-specific configurations. By combining improved spectral encodings, lightweight architecture, and robust padding-band handling, Fast-SEnSeI improves efficiency and accuracy on embedded satellite hardware. Experiments on Landsat 8 and Sentinel-2 datasets show that Fast-SEnSeI maintains strong performance across a wide range of band combinations, a capability not achievable with traditional fixed-band models. Compared to SEnSeI-v2, it delivers higher throughput and a smaller memory footprint without loss of accuracy. Fast-SEnSeI processes a 512×512 image with five spectral bands in 30 ms on an ARM Cortex-A53, demonstrating suitability for real-time on-board use. The module is generic and can support any downstream model, not only the cloud segmentation network presented in this work.

For future work, an important step is evaluating Fast-SEnSeI directly on raw sensor data, which will be the actual input available on board. Current preprocessing pipelines for Level-1C Sentinel-2 and Level-1T Landsat 8 apply radiometric and geometric corrections, many of which are too computationally expensive to perform in orbit. Since manually annotated cloud masks are not available for raw data, this could be approached by simulating raw sensor measurements from the higher-level products.

\section{Acknowledgments}
This work was performed under the Investing in Industrial Innovation Public Private Partnership co-funding programme run by the ESA \(\Phi\)-lab, under ESA Contract No.\ 4000142866/23/I-DT-lr. The views expressed herein can in no way be taken to reflect the official opinion of the European Space Agency.

\section{Preprint notice}
This is a preprint of a paper accepted for the \href{https://atpi.eventsair.com/edhpc-2025/}{EDHPC 2025 Conference}.

\bibliographystyle{IEEEtran}
\bibliography{refs}

\end{document}